\documentclass[conference]{IEEEtran}
\IEEEoverridecommandlockouts

\usepackage{times}

\usepackage[numbers]{natbib}
\usepackage{multicol}

\usepackage{graphics,graphicx,caption,float,subcaption,booktabs,xcolor,multirow,array,color,ifthen,tabu,colortbl,dblfloatfix,url,xparse,mathtools,patchcmd,algorithm,algorithmic,amssymb,xspace,nicefrac,microtype,amsmath,amsfonts,bm,ragged2e,tikz,stackengine,etoolbox,xpatch,enumerate,xstring,setspace,tabularx,makecell,changepage,cuted,titlesec,wrapfig,tcolorbox, sidecap, xspace, dsfont, algorithm, listings}

\usepackage[pagebackref=true,breaklinks=true,colorlinks=true,bookmarks=false,citecolor=orange,linkcolor=orange,urlcolor=teal]{hyperref}

\newcommand{\ours}{HRP\xspace}

\newcommand{\para}[1]{\noindent \textbf{#1}: }

\newcommand{\website}{\url{https://hrp-robot.github.io}}

\begin{document}

\title{HRP: Human Affordances for Robotic Pre-Training}

\author{ \hspace{-1.5em} Mohan Kumar Srirama \hspace{1em} Sudeep Dasari$^\psi$\thanks{$\psi$ Denotes equal advising.} \hspace{1em} Shikhar Bahl$^\psi$ \hspace{1em} Abhinav Gupta$^\psi$ \\ \\
Carnegie Mellon University
}

\maketitle

\begin{abstract}
In order to \textit{generalize} to various tasks in the wild, robotic agents will need a suitable representation (i.e., vision network) that enables the robot to predict optimal actions given high dimensional vision inputs. However, learning such a representation requires an extreme amount of diverse training data, which is prohibitively expensive to collect on a real robot. How can we overcome this problem?  Instead of collecting more robot data, this paper proposes using internet-scale, human videos to extract ``affordances," both at the environment and agent level, and distill them into a pre-trained representation. We present a simple framework for pre-training representations on hand, object, and contact ``affordance labels" that highlight relevant objects in images and how to interact with them. These affordances are automatically extracted from human video data (with the help of off-the-shelf computer vision modules) and used to fine-tune existing representations. Our approach can efficiently fine-tune \textit{any} existing representation, and results in models with stronger downstream robotic performance across the board. We experimentally demonstrate (using 3000+ robot trials) that this affordance pre-training scheme boosts performance by a minimum of 15$\%$ on 5 real-world tasks, which consider three diverse robot morphologies (including a dexterous hand). Unlike prior works in the space, these representations improve performance across 3 different camera views. Quantitatively, we find that our approach leads to higher levels of generalization in out-of-distribution settings. For code, weights, and data check: \website

\end{abstract}

\IEEEpeerreviewmaketitle

\section{Introduction} 
\label{sec:introduction}

\begin{figure}[t!]
     \centering
     \includegraphics[width=\linewidth]{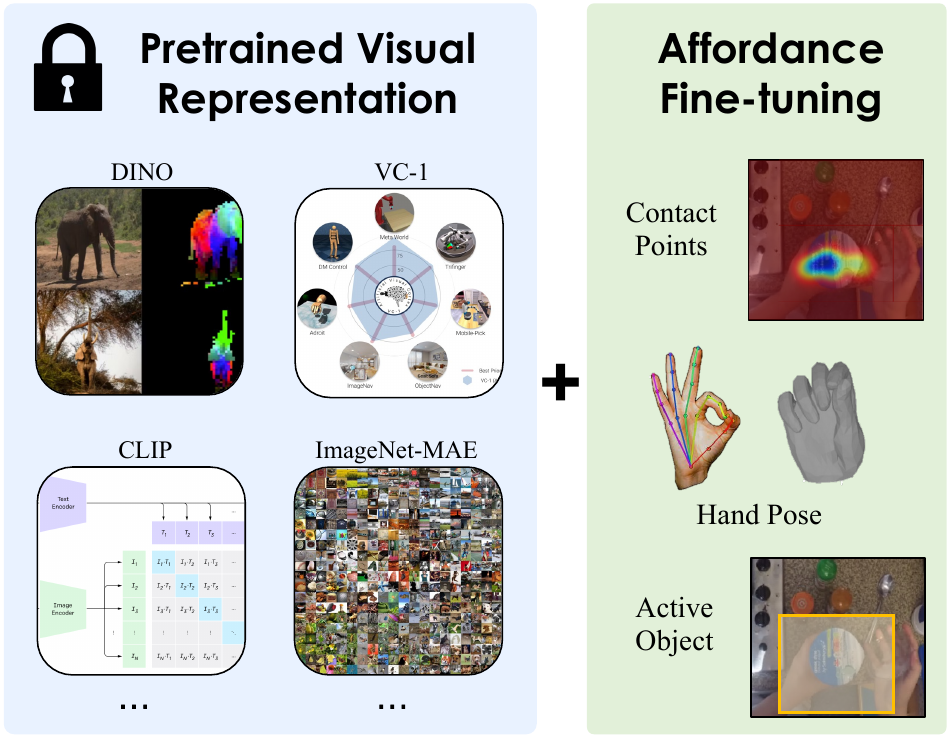}
     \caption{ Pre-trained representations offer a scalable solution to the robotics data bottleneck~\cite{r3m,realmvp,vc1}, but existing methods fail to reliably improve over simple baselines like ImageNet~\cite{dasari2023datasets,Burns2023WhatMakesPVR}. Thus, we present \textbf{HRP}, a method that mines affordances (e.g., contact, hand pose, and object labels) from human videos and uses them to improve self-supervised visual encoders. Our best \ours representation consistently outperforms 6 SOTA baselines by $\geq \textbf{20\%}$ across 5 diverse tasks and 3 camera views.}
     \label{fig:teaser}

    \vspace{-0.5cm}
 \end{figure}

A truly generalist robotic agent must acquire diverse manipulation skills (ranging from block stacking to pouring) that work with novel objects and remain robust to realistic environmental disturbances (e.g., lighting changes, small camera shifts). Due to the scale of this challenge, the field has trended towards learning these agents directly from data~\cite{levine2016end,pinto2016supersizing}, particularly robot trajectories collected either by expert demonstrators or autonomously by the agents themselves (via Reinforcement Learning~\cite{sutton2018reinforcement}). Unfortunately, there are innumerable objects/environments, so roboticists cannot tractably collect enough real-world demonstration data and/or design a simulator that captures all this diversity. 

One promising solution for this ``data challenge" is for the robot to learn a \textit{suitable representation} from Out-Of-Domain (OOD) data that can be transferred into the robotics domain. For example, prior work~\cite{r3m,realmvp,vc1} trained self-supervised image encoders on large scale datasets of human videos (e.g., Ego4D~\cite{grauman2022ego4d}), using standard reconstruction objectives and contrastive learning~\cite{cpc} objectives -- e.g., Masked Auto-Encoders~\cite{mae} (MAE) and Temporal Contrastive Networks~\cite{tcn} (TCN) respectively -- developed by the broader learning community. After pre-training, these representations are used to initialize downstream imitation learning~\cite{schaal1997learning} algorithms. This formula is extremely flexible, and can substantially reduce the amount of robot data required for policy learning. However, the representations are often only effective when using specific camera views and robot setups. Furthermore, independent evaluations~\cite{dasari2023datasets,Burns2023WhatMakesPVR} recently showed that these representations cannot improve (on average) over the most obvious baseline -- a self-supervised ImageNet representation~\cite{mae,imagenet}! 

\begin{figure*}[t!]
     \centering
     \includegraphics[width=0.95\linewidth]{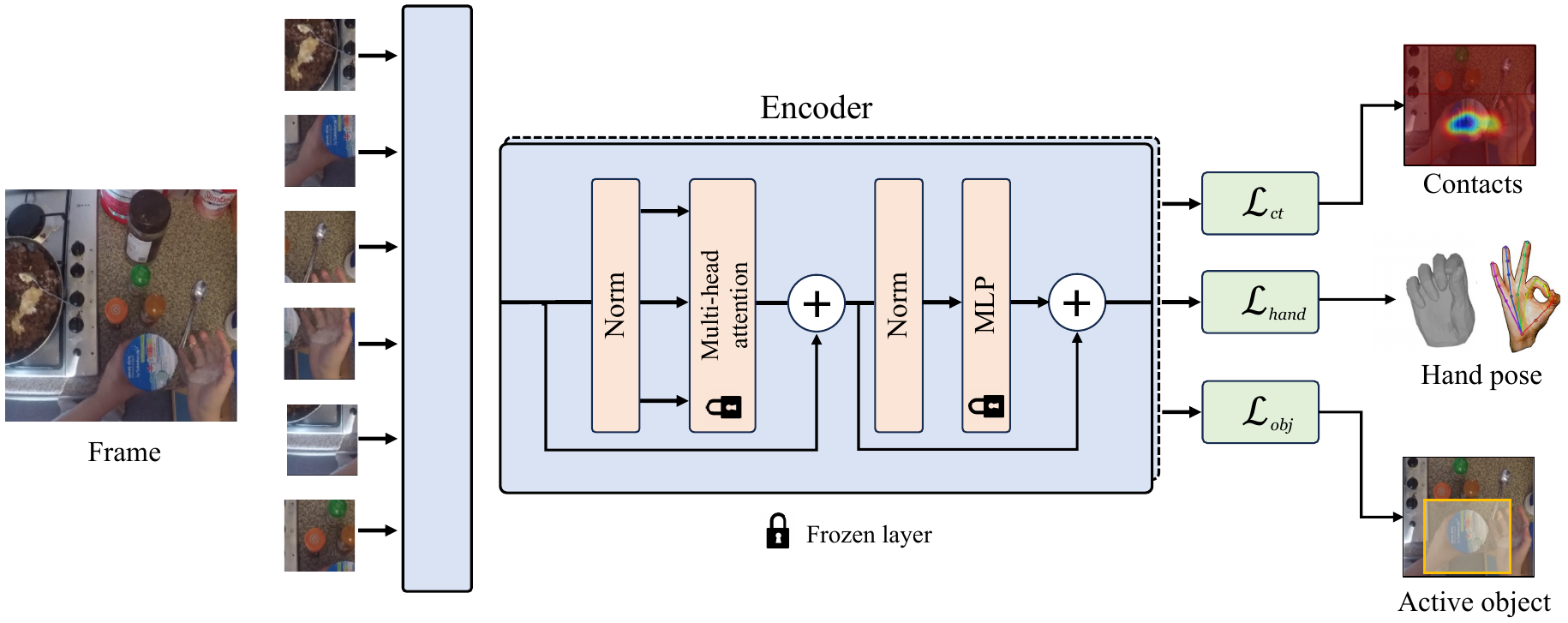}
     \caption{ \ours fine-tunes a pre-trained encoder to predict three classes of human affordance labels via L2 regression. Specifically, the network must predict future contact points, human hand poses, and the target object given an input frame from the video stream. These affordance labels are mined autonomously from a human video dataset~\cite{grauman2022ego4d} using off-the-shelf vision detectors~\cite{100doh}. \ours representations are then fine-tuned to solve downstream manipulation tasks via behavior cloning. }
     \label{fig:method}
     \vspace{-0.5cm}
 \end{figure*}

This result is surprising since robot trajectories and human video sequences share so much common structure: both modalities contain an agent (e.g., human or robot) using their end-effector (e.g., human hand, robot gripper) to manipulate objects in their environment. Ideally, representations trained on this data would learn useful object attributes (e.g., where to grasp a mug), and spatial relationships between the end-effector and target objects. We hypothesize that traditional self-supervised learning objectives are unable to extract this information from human video data, and that explicitly predicting these object/spatial features would result in a stronger robotic representation (i.e., higher down-stream control performance). Our key insight is that abandoning self-supervision comes at minimal cost -- the necessary object and hand labels can be scalably mined using off-the-shelf vision pipelines.

This paper proposes Human affordances for Robotic Pre-training (\ours),  a semi-supervised pipeline to learn effective robotic representations from human video. \ours works in two stages: first, it extracts hand-object ``affordance" information -- i.e., which objects in the scene are graspable and how the robot should approach them -- from human videos using off-the-shelf tracking models~\cite{100doh,FrankMocap_2021_ICCV}. These affordances are then distilled into a pre-existing representation network (e.g., ImageNet MAE~\cite{mae}), \textit{before} the policy fine-tuning stage. This paradigm allows us to inject useful information into the vision encoder, while preserving the flexibility of self-supervised pre-training -- i.e., all labels are automatically generated and the network can be easily slotted into downstream robotic policies/controllers via fine-tuning.  To summarize, \textbf{we learn stronger robotic representations by predicting object interactions and hand motion from human video dataset images} (see Fig.~\ref{fig:teaser}). Our investigations and experiments lead to the following contributions: 

\begin{enumerate}
    \item We present a semi-supervised learning algorithm -- \ours -- that leverages off-the-shelf human affordance models to learn effective robotic representations from human video. The proposed pipeline strongly outperforms representations learned purely via self-supervision.
    \item Applying \ours to 6 pre-existing representations (including ImageNet \cite{imagenet, mae}, VC-1 \cite{vc1}, and DINO \cite{dino}) substantially boosts  robot performance. This conclusion is backed by \textbf{3000+ robot trials}, and replicates across 3 camera views, 3 distinct robotic setups, and 5 manipulation tasks!
    \item Our ablation study reveals that \ours's three affordance objectives (hand, object, and contact based loss terms) are all critical for effective representation learning.
    
    \item We show that \ours representations generalize across different imitation learning stacks -- \ours improves diffusion policy~\cite{chi2023diffusionpolicy} performance by $20\%$!
    \item Our best representation, which increases performance by 20$\%$ over State-of-the-Art (SOTA), will be fully open-sourced, along with all code and data.
\end{enumerate}

\section{Related Work} 
\label{sec:relatedwork}

\para{Representation Learning in Robotics} End-to-end policy learning offers a scalable formula for acquiring robotic representations: instead of hand-designing object detectors or image features, a visual encoder is directly optimized to solve a downstream robotic task~\cite{levine2016end}. Numerous works applied this idea to diverse tasks including bin-picking~\cite{kalashnikov2018qt,levine2018learning,pinto2016supersizing}, in-the-wild grasping~\cite{gupta2018robot,song2020grasping}, insertion~\cite{dasari2021rb2,levine2016end}, pick-place~\cite{rt12022arxiv}, and (non-manipulation tasks like) self-driving~\cite{bojarski2016end,pomerleau1988alvinn,chen2020learning}. Furthermore, secondary learning objectives -- e.g., dynamics modeling~\cite{hafner2020dream,whitney2019dynamics}, observation reconstruction~\cite{nair2018visual}, inverse modeling~\cite{dasari2021transformers}, etc. -- can be easily added to improve data efficiency. While this paradigm can be effective, learning purely from robot data requires an expensive data collection effort (e.g., using an arm farm~\cite{levine2018learning,kalashnikov2018qt}, large-scale tele-operation~\cite{rt12022arxiv}, or multi-institution data collection~\cite{dasari2019robonet,open_x_embodiment_rt_x_2023}), which is infeasible for (most) task settings. 

To increase data efficiency, prior work applied self-supervised representation learning algorithms on out-of-domain datasets (like Ego4D~\cite{grauman2022ego4d}), and then fine-tuned the resulting representations to solve downstream tasks with a small amount of robot data -- e.g., via behavior cloning on $\leq 50$ expert demonstrations~\cite{r3m,vc1,realmvp}, directly using them as a cost/distance function to infer robot actions~\cite{vip,wang2023manipulate}, or directly pre-training robot policies from extracted human actions. \cite{shaw2022video, dexvip, kannan2023deft}. While this transfer learning paradigm can certainly be effective, it is unclear if these robotic representations~\cite{vc1,r3m,realmvp} provide a substantial boost over pre-existing vision baselines~\cite{dasari2023datasets,Burns2023WhatMakesPVR}, like ImageNet MAE~\cite{mae} or DINO~\cite{dino}. One potential issue is that roboticists often use the same exact pre-training methods from the vision community, but merely apply them to a different data mix (e.g., VC-1~\cite{vc1} applies MAE~\cite{mae} to Ego4D~\cite{grauman2022ego4d}). Thus, the resulting representations are never forced to key in on object/agent level information in the scene. This paper proposes a simple formula for injecting this information into a vision encoder, using a mix of hand and object affordance losses, which empirically boost performance on robotic tasks by 25\%. \\

\para{Affordances from Humans} 
\ours is heavily inspired by the \textit{affordance learning} literature in computer vision~\cite{gibson1979ecological, gibson1966senses}. These works use human data as a probe to learn environmental cues (i.e., affordances) that tell us how humans might interact with different objects. These include physical~\cite{eigen2014predicting, bansal2016marr, gupta20113d, zhu2016inferring, hassanin2018visual, zhao2013scene, myers2015affordance} and/or semantic~\cite{roy2016multi, sawatzky2017weakly} scene properties, or forecast future poses \cite{koppula2015anticipating, rhinehart2016learning, gao2017red,jain2016recurrent,huang2014action,vondrick2016predicting,abu2018will,lan2014hierarchical,villegas2017decomposing,grauman2022ego4d, furnari2020rolling, mascaro2022intention, girdhar2021anticipative}  Affordances can also be learned at object or part levels~\cite{zhu2014reasoning,furnari2017next, hap, hotspots, hoi, ye2023affordance}. Usually such approaches leverage human video datasets~\cite{grauman2022ego4d, EPICKITCHENS, youcook, Damen2018EPICKITCHENS} or use manually annotated interaction data~\cite{liu2022hoi4d,darkhalil2022epic, 100doh}. In addition to these cues, robotic affordances must consider how to move before and after interaction~\cite{bahl2023affordances, hmr}. A simple, scalable way to capture this information is by detecting these cues from human hand poses in monocular video streams~\cite{wang2020rgb2hands, hmr, FrankMocap_2021_ICCV, lugaresi2019mediapipe}, which show robots reaching for and manipulating diverse, target objects. 
Our method combines these three approaches to create a human affordance dataset automatically from human video streams. The labels generated during this process are distilled into a representation and used to improve downstream robotics task performance.

\begin{figure}[t]
     \centering
     \includegraphics[width=0.95\linewidth]{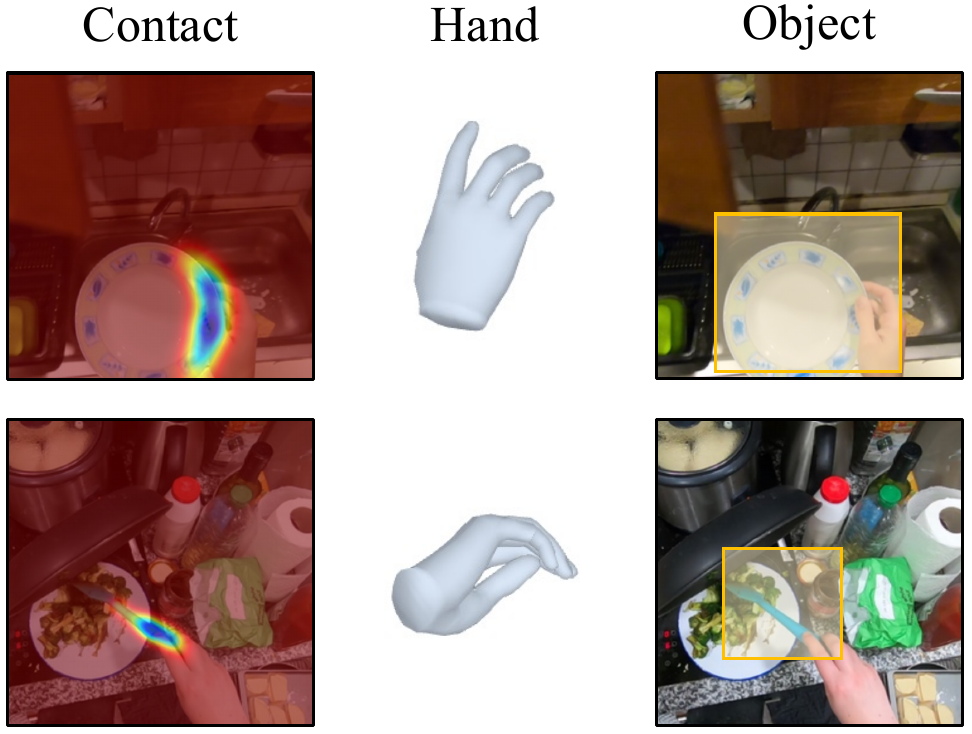}
     \caption{ We extract 3 affordances -- contact heatmaps, hand poses and active object bounding boxes -- from human videos. }
     \label{fig:affordances}
     \vspace{-0.5cm}
 \end{figure}

\section{Preliminaries} 
\label{sec:prelims}

\subsection{Visual Representation Learning} 
\label{sec:prelims-repr}

\begin{figure*}[t]
    \vspace{-0.4in}
    \centering
    \includegraphics[width=\linewidth]{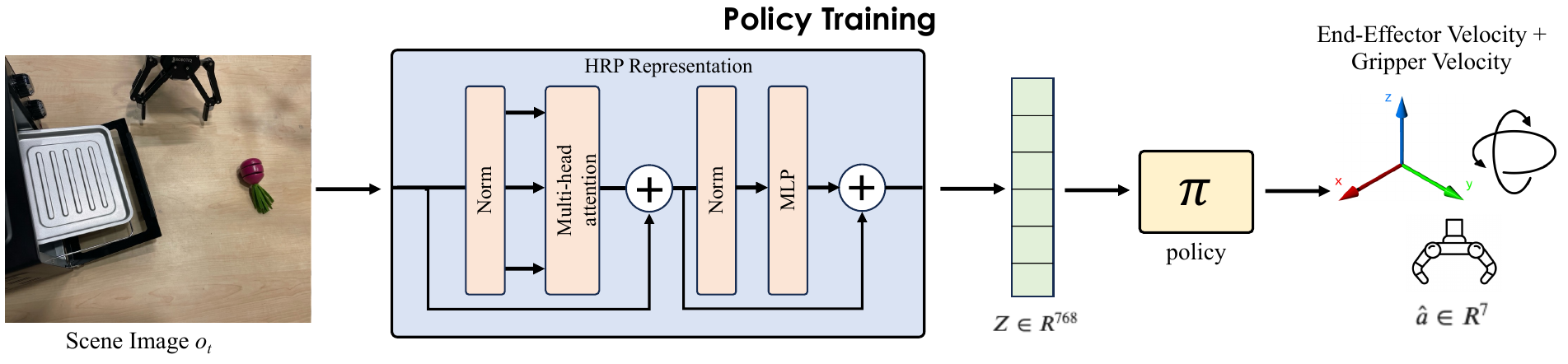}
    \caption{ \small We present our policy training pipeline, which uses Behavior Cloning (BC) to train policy $\pi$, using optimal expert demonstrations. The image observation ($o_t$) is processed using our \ours representations resulting in a latent vector $z$. The policy uses $z$ to predict end-effector velocity actions (delta ee-pose/gripper), which are directly executed on the robot during test-time. }
    \label{fig:policy-training}
    \vspace{-0.5cm}
\end{figure*}

Our goal is to learn a visual encoder network $f_\theta$ that takes an input image $I$ and processes it into a low-dimensional vector $f_\theta(I) \in \mathcal{R}^d$. This resulting ``embedding vector" would ideally encode important scene details for robotic policy learning -- like the number and type of objects in a scene and their relationship to the robot end-effector. In this paper, $f_\theta$ is a transformer network (specifically ViT-B~\cite{vit}, with patch size 16 and $d=768$) parameterized with network weights $\theta$. But to be clear, all our methods are network architecture agnostic.\\

\para{Self-Supervised Learning} The computer vision community has broadly adopted \textit{self-supervised} representation learning algorithms that can pre-train network weights without using \textit{any} task-specific supervision. This can be accomplished using a \textit{generative learning objective}~\cite{vae}, which trains $f_\theta$ alongside a decoder network $D$ that reconstructs the original input image input from the representation. Another common approach is \textit{contrastive learning}~\cite{cpc,moco}, which optimizes $f_\theta$ to maximize the mutual information between the encoding and the input image (i.e., place ``similar" images closer in embedding space). In practice, these methods can learn highly useful features for downstream vision tasks~\cite{mae,moco}, but struggle in robotics settings~\cite{dasari2023datasets,Burns2023WhatMakesPVR}. Our goal is to inject these features into an existing self-supervised network, with an affordance-driven fine-tuning stage.

\subsection{Extracting Affordance Labels from Human Data}
\label{sec:prelims-affordances}

Before we can do any fine-tuning, we must first curate a suitable human affordance dataset $\mathcal{D}_H$. Thankfully this task can be done automatically using off-the-shelf vision modules, applied to a set of $150K$ human-object interaction videos from Ego4D (originally sampled by \citet{r3m}). These are subsets of larger videos (around 1.2K) videos, which were further broken down into shorter clips. Each clip contains a semantically meaningful action by the human. Each video clip $V$ contains image frames $V = \{I_1, \dots, I_T\}$ that depict human hands performing tasks and moving around in the scene. From these images, we obtain \textbf{contact locations}, \textbf{future hand p-oses}, and \textbf{active object labels} (examples in Fig.~\ref{fig:affordances}) that capture various agent-centric properties (how to move and interact) and environment centric properties (where to interact) at multiple scales, i.e. contact-level and object-level. The following sections detail how each of these labels were generated.\\

\para{Contact Locations} To extract contact locations for an image $I_t$ (with no object contact), we find the frame $I_j; j > t$ where contact with a given object will begin, using a hand-object interaction detection model~\cite{100doh}. Then, we use $I_j$ to find the active object $O_j$ and the hand mask $M_j$. The points intersecting $M_j$ and $O_j$ (acquired via skin segmentation) are our contact affordances ($C_j$). To account for motion between $I_t$ and $I_j$, we compute the homography matrix between the frames and project those points forward. This is done using standard SIFT feature tracking~\cite{zhou2009object}: $C_t = H_{j,t} C_j$. \textit{In other words, the contact locations denote where in $I_t$ the human will contact in the future.} Note that there could be a different number of points for each contact scenario, which is non-ideal for learning. Thus, we fit a Gaussian Mixture Model with $k = 5$ modes on $C_t$ to make a uniform contact descriptor -- defined as the means $c_t$ of the mixture model. For more details on extraction, we refer to Appendix~\ref{app:data_pipeline}.\\

\para{Future Hand Poses} This affordance label captures how the human moves next (e.g., to complete a task or reach an object), as the video $V$ progresses. Given a current frame $I_t$, we detect the human hand's 2d wrist position ($h_{t+k}$) in a future frame $I_{t+k}$, where usually $k=30$ (empirically determined). This is done using the Frank Mocap~\cite{FrankMocap_2021_ICCV} hand detector. To correctly account for the human's motion, these wrist points are back-projected (again using the camera homography matrix) to $I_t$ to create the final ``future wrist label," $h_t = H_{t+k,t} h_{t+k}$.\\

\para{Active Object Labels} In a similar manner to the contact location extraction, we run a hand-object interaction detection model~\cite{100doh} on $V$ to find the image where contact began $I_c$. The same detector is used to find the four bounding box coordinates of the object that is being interacted with, which we refer to as the ``active object." These coordinates $b_c$ are then projected to every other frame $I_t$, using the homography matrix (see above). This results in an active bounding box $b_t$ for each image in $V$.

   \begin{figure*}[t!]
     \centering
     \includegraphics[width=\linewidth]{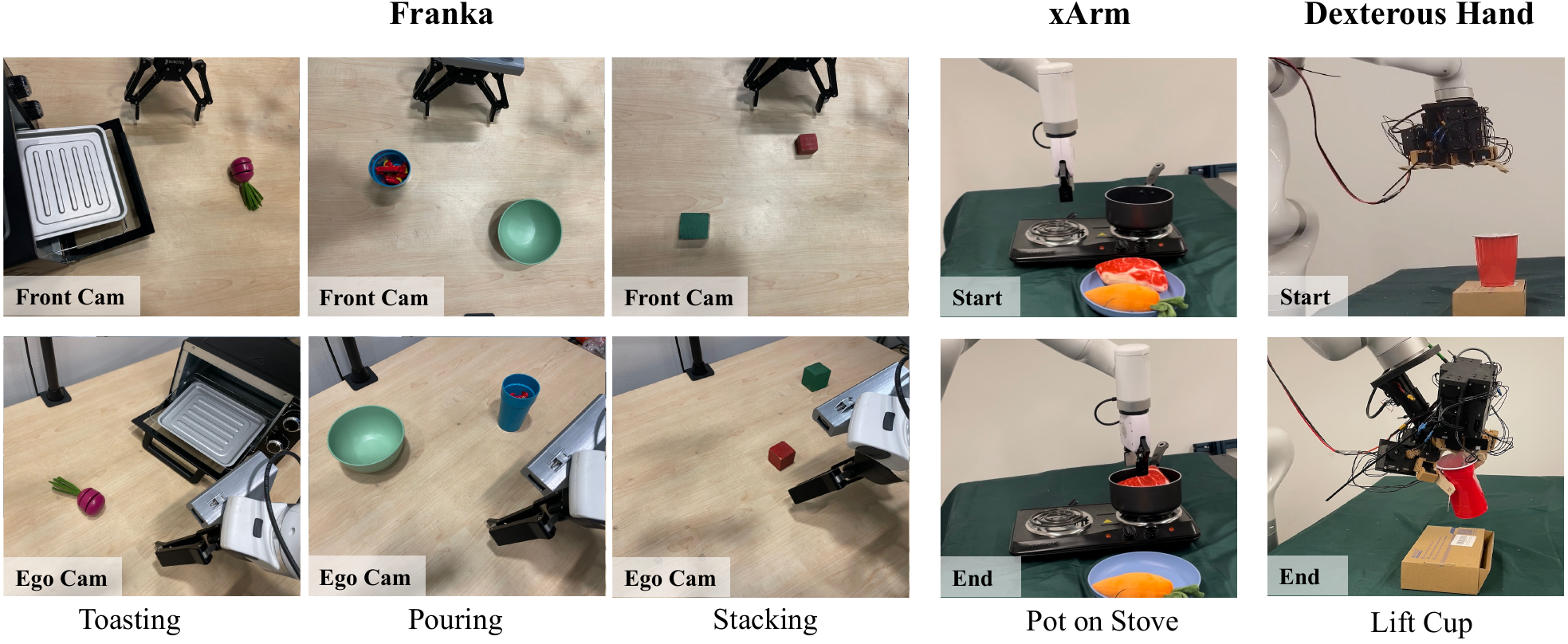}
     \caption{ Our experiments consider 5 unique manipulation tasks, ranging from classic block-stacking to a multi-stage toasting scenario. These tasks are implemented on 3 unique robot setups, including a high Degree-of-Freedom dexterous hand (right). The 3 camera views shown -- front, ego, and side views (for xArm/dexterous hand) -- are the same views ingested by the policy during test-time. Note that 3 of the tasks consider 2 unique camera views in order to test for robustness!}
     
     \label{fig:tasks}
     \vspace{-0.5cm}
 \end{figure*}

\section{Introducing \ours} 
\label{sec:methods}

A variety of visual pre-training tasks have been shown to help with downstream robotic performance-- ranging from simple ImageNet classification~\cite{shah2021rrl} to self-supervised learning on human video~\cite{realmvp, r3m, vip, vc1, pari2021surprising}. Although these approaches operate on human videos and simple image frames, they fail to explicitly model the rich hand-object contacts depicted. In contrast, we believe explicitly modeling the \textit{affordances}~\cite{gibson1966senses} in this data could allow us to learn useful information about the agent's intents, goals, and actions. Indeed, past work has shown that affordances can act as strong prior for manipulation~\cite{zhou2023hacman, ju2024robo, shridhar2022cliport, zeng2020transporter, bahl2023affordances, chang2024look, huang2023voxposer, bharadhwaj2023towards} in general. Moreover, this information can be represented in many different formats, such as physical attributes, geometric properties, interactions, object bounding boxes, or motion forecasting. We observe that most tasks of interests humans perform are with their hands. We thus focus on training our model to predict hand-object interactions and hand motion.

We present \ours, a simple and effective representation learning approach that injects hand-object interaction priors into a self-supervised network, $f_\theta$, using an automatically generated human affordance dataset, $\mathcal{D}_H$ (see above for definitions and dataset mining approach). \ours is illustrated in Fig.~\ref{fig:method}, and the following sections describe its implementation in detail.

\subsection{Training \ours}
\label{sec:methods-reprs}

The initial network $f_\theta$ is fine-tuned using batches sampled from the human dataset: $(I_t, c_t, h_t, b_t) \sim \mathcal{D}_H$, where $c_t$, $h_t$, and $b_t$ are contact, hand, and object affordances corresponding to image $I_t$ (see Sec.~\ref{sec:prelims-affordances} for definitions). Some frames may not include all 3 affordances, so we include 3 mask variables -- $m^{(c)}_t, m^{(h)}_t, m^{(b)}_t$ -- so the missing values can be ignored during training. We add 3 small affordance modules -- $p_{c}$, $p_h$, $p_b$ -- on top of $f_\theta$ that are trained to regress the respective affordances for $I_t$. This results in the following three loss functions:

\begin{equation}
    \mathcal{L}_\text{ct} =||c_{t} - p_c(f_\theta(I_t))||_2
    \label{eq:loss-ct}
\end{equation}
\begin{equation}
    \mathcal{L}_\text{hand} =||h_{t} - p_h(f_\theta(I_t))||_2
    \label{eq:loss-hand}
\end{equation}
\begin{equation}
    \mathcal{L}_\text{obj} =||b_{t} - p_b(f_\theta(I_t))||_2
    \label{eq:loss-obj}
\end{equation}

The full loss is: 

\begin{equation}
    \mathcal{L} = m^{(c)}_t \lambda_\text{ct}\mathcal{L}_\text{ct} +  m^{(h)}_t\lambda_\text{hand}\mathcal{L}_\text{hand} +
    m^{(b)}_t\lambda_\text{obj}\mathcal{L}_\text{obj} 
    \label{eq:loss}
\end{equation}
Where the $\lambda$s are hyper-parameters that control the relative weight of each affordance loss. We empirically found $\lambda_\text{obj}=0.05, \lambda_\text{ct}=0.005, \lambda_\text{hand}=0.5$ to be optimal for downstream performance (see Appendix~\ref{app:hparam}).

\subsection{Implementation Details}
\label{sec:methods-implement}

Our affordance dataset ($\mathcal{D}_H$) is at least an order of magnitude smaller than the pre-training image dataset initially used by the baseline representation (e.g., ImageNet has 1M frames v.s. our 150K). To preserve the useful features learned from the larger pre-training distribution, we keep most of the parameters in $\theta$ fixed during \ours fine-tuning. Specifically, we only fine-tune the baseline network's normalization layers and leave the rest fixed, which has been shown to be an effective approach~\cite{giannou2023expressive, zhao2023tuning}. In the case of our ViT-B this amounts to fine-tuning only the LayerNorm parameters $\gamma$ and $\beta$:
\begin{equation}
    \text{LayerNorm}(x) = \frac{x - \mu}{\sigma}\gamma + \beta 
\end{equation}

These parameters are fine-tuned to minimize $\mathcal{L}$ using standard back-propagation and the ADAM~\cite{kingma2014adam} optimizer.

\section{Experimental Details}
\label{sec:expsetup}

Our contributions are validated using a simple empirical formula: first, \ours is applied to each baseline model (listed below). Then, (following standard practice~\cite{r3m,vc1,dasari2023datasets}) the resulting representation is fine-tuned into a manipulation policy using behavior cloning. Details for each stage are provided below, and the \ours is illustrated in Fig.~\ref{fig:method}. \\

\para{Baseline Representations} We chose 6 representative, SOTA baselines from both the vision and robotics communities: 
\begin{enumerate}
    \item \textbf{ImageNet MAE} was pre-trained by applying the Masked Auto-Encoders~\cite{mae} (MAE) algorithm to the ImageNet-1M dataset~\cite{imagenet}. It achieved SOTA performance across a suite of vision tasks, and is the first self-supervised representation to beat supervised pre-training. We use the standard Masked Auto Encoder training scheme for this, using hyperparmaeters from MAE \cite{mae}.
    \item \textbf{Ego4D MAE} was trained by applying the MAE algorithm to a set of 1M frames sampled from the Ego4D dataset~\cite{grauman2022ego4d}. For consistency with prior work, we use the same 1M frame-set sampled by the R3M authors~\cite{r3m}. We use the standard Masked Auto Encoder training scheme for this, using hyperparmaeters from MAE \cite{mae}.
    \item \textbf{CLIP~\cite{Clip}} is a SOTA representation for internet data. It was learned by applying contrastive learning~\cite{cpc} to a large set natural language - image pairs crawled from internet captions. We used publicly available model weights.
    \item \textbf{DINO~\cite{dino}} was trained using a self-distillation algorithm that encourages the network to learn local-to-global image correspondences. DINO's emergent segmentation capabilities could be well suited for robotics, and it has already shown SOTA performance in sim~\cite{Burns2023WhatMakesPVR}. We used publicly available model weights.
    \item \textbf{MVP~\cite{realmvp}} was trained by applying MAEs to a mix of in-the-wild datasets (100 DoH~\cite{100doh}, Ego4D~\cite{grauman2022ego4d}, etc.). The authors showed strong performance on various manipulation tasks. We used publicly available model weights.
    \item \textbf{VC-1~\cite{vc1}}  was trained in a similar fashion to MVP, but used a larger dataset mix. It showed strong performance on visual navigation tasks. We used publicly available model weights.
\end{enumerate}
Note that each baseline is parameterized with the same ViT-B encoder w/ patch size 16 (see Sec.~\ref{sec:prelims-affordances}), to ensure apples-to-apples comparisons.\\

\begin{figure*}[t]
    \vspace{0.1in}
    \centering
    \begin{subfigure}[b]{0.47\linewidth}
        \centering
        \includegraphics[width=\linewidth]{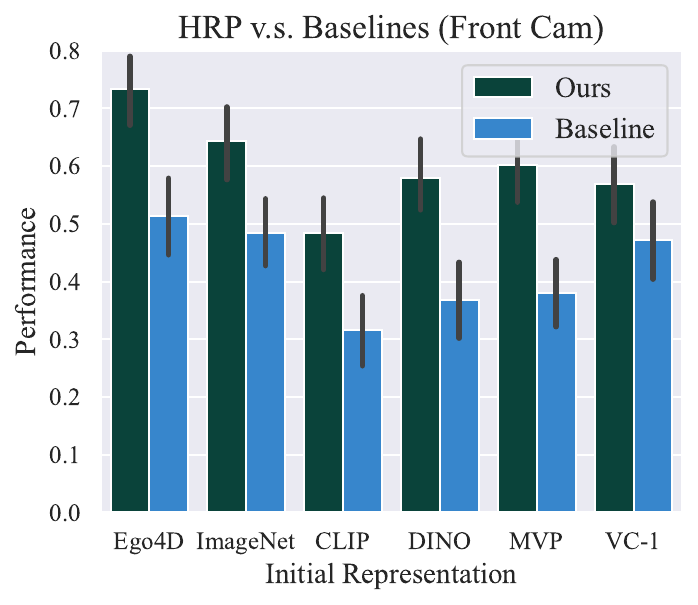}
        \label{fig:main-frontcam}
    \end{subfigure}
    \begin{subfigure}[b]{0.47\linewidth}
        \centering
        \includegraphics[width=\linewidth]{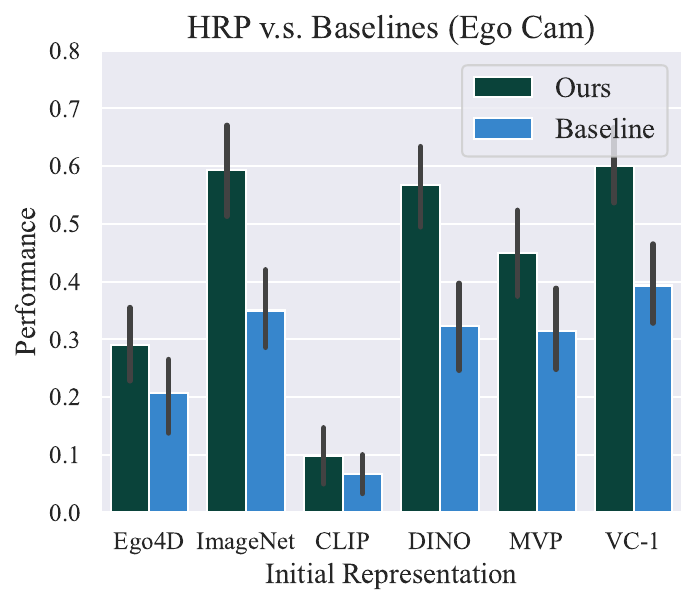}
        \label{fig:main-egocam}
    \end{subfigure}
    \vspace{-0.5cm}
    \caption{ \small We apply \ours to 6 different baseline representations and plot how it affects performance on average across the \textit{toasting}, \textit{pouring}, and \textit{stacking} tasks. This evaluation procedure is repeated using two distinct cameras (shown in Fig.~\ref{fig:tasks}) in order to test if \ours representation are robust to view shifts. We find that \ours representations consistently and substantially outperform their vanilla baselines, and that this effect holds across both the front (left) and ego (right) cameras. In fact, our strongest representation -- ImageNet + \ours -- delivers SOTA performance on both views! }
    \vspace{-0.5cm}
    \label{fig:main}
\end{figure*}

\para{Policy Learning} Each representation is evaluated on downstream robotic manipulation tasks, by fine-tuning it into a policy ($\pi$) using Behavior Cloning~\cite{pomerleau1988alvinn,schaal1999imitation,ross2011reduction}. Note that $\pi$ must predict the expert action ($a_t$ -- robot motor command) given the observation ($o_t$ -- input image and robot state): $a_t \sim \pi (\cdot | o_t)$. And $\pi$ is learned using a set of \textit{50 expert demonstrations} $\mathcal{D} = \{ \tau_1, \dots, \tau_{50}\}$, where each demonstration $\tau_i = [(o_0, a_0), \dots, (o_T, a_T)]$ is a trajectory of expert observation-action tuples. In our case, $\pi$ is parameterized by a small 2-layer MLP ($p$) placed atop the pre-trained encoder $p(f(o_t))$ that predicts a Gaussian Mixture policy distribution w/ 5 modes. Both the policy network and visual encoder are optimized end-to-end (using ADAM~\cite{kingma2014adam} w/ $lr=0.0001$ for 50K steps) to maximize the log-likelihood of expert actions: $\text{max}_{p,f} log(\pi(a_t | p(f(o_t))))$. During test time actions are sampled from this distribution and executed on the robot: $a_t \sim \pi(\cdot | p(f(o_t)))$. This is a standard evaluation formula that closely follows best practices from prior robotic representation learning work~\cite{robomimic,dasari2023datasets}. \\

\para{Real World Tasks} We fine-tune policies for each representation on the 5 diverse tasks listed below, which are implemented on 3 unique robotic setups, including a dexterous hand (illustrated in Fig.~\ref{fig:tasks}). 50 expert fine-tuning demonstrations were collected for each task via expert tele-operation. Note that the stacking, pouring, and toasting tasks were \textit{evaluated twice using different camera views} to test robustness!

\begin{itemize}
    \item \textbf{Stacking:} The stacking task requires the robot to pick up the red block and place it on the green block. During test time both blocks' starting positions are randomized to novel locations (not seen in training). A trial is marked as successful if the robot correctly picks and stacks the red block, and half successful if the red block is unstably placed on the green block. This task is implemented on a Franka robot and uses both an Ego and Front camera viewpoint. 

    \item \textbf{Pouring:} The pouring task requires the robot to pick up the cup and pour the material (5 candies) into the target bowl. During test time we use novel cups and bowls and place each in new test locations. This task's success metric is the fraction of candies successfully poured (e.g., $2 / 5$ candies poured $\rightarrow 0.4$ success). This task was also implemented on the Franka using Ego and Front cameras.

    \item \textbf{Toasting:} The toasting task requires the robot to pick up a target object, place it in the toaster oven, and shut the toaster. This is a challenging, multi-stage task. During test time the object type, and object/toaster positions are both varied. A test trial is marked as successful if the whole task is completed, and 0.5 successful if the robot only successfully places the object. This is the final task implemented on Franka w/ Ego and Front camera views.

    \item \textbf{Pot on Stove:} The stove task requires picking up a piece of meat or carrot from a plate and placing it within a pot on a stove. During test time, novel ``food" objects are used and the location is randomized. A trial is marked as successful if the food is correctly placed in the pot. This task is implemented on a xArm and uses the side camera view.

    \item \textbf{Hand Lift Cup} This task requires a dexterous hand to reach, grasp, and lift up a deformable red solo up. The hand's high dimensional action space ($\mathcal{R}^{20
    }$) makes this task especially challenging. A trial is marked successful if the cup is stably grasped and picked. This task is implemented on a custom dexterous hand using a side camera view.
\end{itemize}

\section{Results} 
\label{sec:results}

Our experiments are designed to answer the following:
\begin{enumerate}

    \item \textbf{Can \ours improve the performance of the pre-trained baseline networks (listed above)}? Does the effect hold across different camera views and/or new robots? (see Sec.~\ref{sec:results-main})
    \item Our affordance labels are generated using off-the-shelf vision modules -- \textbf{does distilling their affordance outputs into a representation (via \ours) work better than simply using those networks as encoders?} (see Sec.~\ref{sec:results-distill})
    \item How does \ours compare against alternate forms of supervision on the same human video dataset? (see Sec.~\ref{sec:results-alternate})
    \item How important are each of the three affordance losses for \ours's final performance? And is it really best to only fine-tune the LayerNorms and leave the other weights fixed? (see Sec.~\ref{sec:results-ablate})
    \item Can \ours handle scenarios with OOD distractor objects during test time? (see Sec.~\ref{sec:results-distract})
    \item Can \ours representations work with different imitation learning pipelines, like diffusion policy~\cite{chi2023diffusionpolicy}? (see Sec.~\ref{sec:results-diffusion})
\end{enumerate}
Note that all experiments were conducted on real robot hardware, and the models were all tested back-to-back (i.e., using proper A/B evaluation) using 50+ trials per model to guarantee statistical significance. Note that all of our figures and tables report success rates (sometimes averaged across the toasting, stacking, and pouring tasks) alongside std. err. to quantify experimental uncertainty -- i.e. $\text{success} \% \pm \text{std. err.}$.

\subsection{Improving Representations w/ \ours}
\label{sec:results-main}

\begin{figure}[t]
    \centering
     \includegraphics[width=\linewidth]{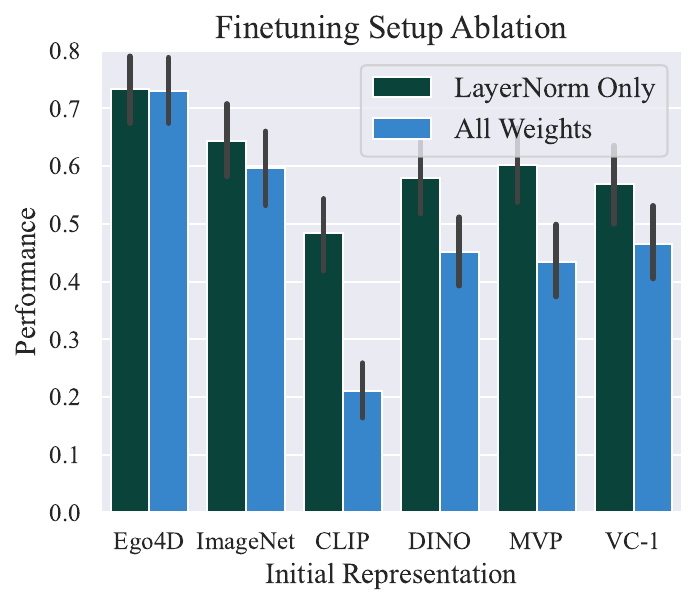}
     \caption{ \small This chart applies an ablated \ours method (full fine-tuning) to the 6 baseline representations, and compares their average performance v.s. standard \ours representations on the \textit{toasting}, \textit{pouring}, and \textit{stacking} tasks (front cam). We find that LayerNorm only fine-tuning is almost always superior.}
     
     \label{fig:ablate-nln}
\end{figure}
\begin{figure}[t]
    \centering
     \includegraphics[width=\linewidth]{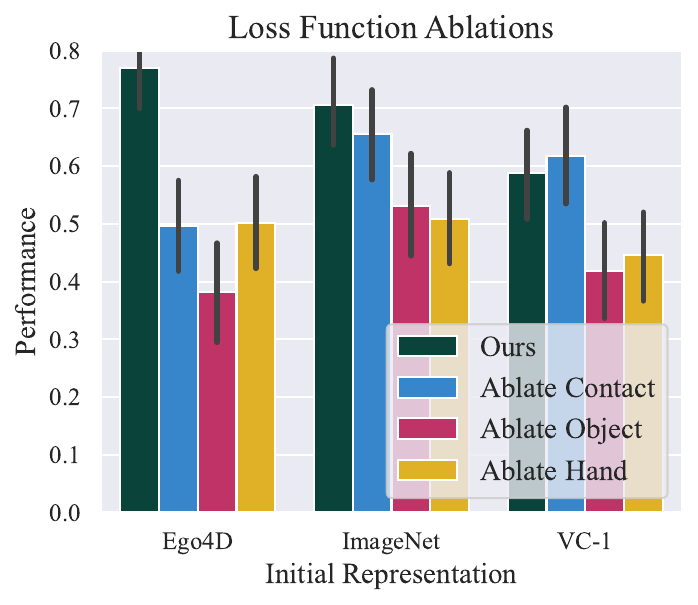}
     \caption{ \small We drop each of the 3 losses in \ours, and compare the ablated method's average performance (across the \textit{toasting}, \textit{pouring}, \textit{stacking} tasks) against full \texttt{\ours} representations. Due to the number of ablations involved, this experiment is only run on the \texttt{Ego4D}, \texttt{ImageNet}, and \texttt{VC-1} base models. We find that the object and hand losses are critical for good performance, but the contact loss only makes a significant impact on the \texttt{Ego4D} base model.}
     \vspace{-0.3cm}
     \label{fig:ablate-loss}
\end{figure}

To begin, we evaluate the 6 baseline representations (detailed in Sec.~\ref{sec:expsetup}) on the \textit{toasting}, \textit{pouring}, and \textit{stacking} tasks using the \textit{front camera view}. Then, we apply \texttt{\ours} to each of these baselines, and evaluate those 6 new models on the same tasks. Average success rates across all 3 tasks are presented in Fig.~\ref{fig:main} (left), and the full table is in the Appendix~\ref{app:front-full-breakdown}. 
First, this experiment demonstrates that ImageNet MAE is still highly competitive on real-world manipulation tasks when compared to other self-supervised representations from the vision~\cite{grauman2022ego4d,dino}, machine learning~\cite{Clip}, and robotics communities~\cite{mvp,vc1}. Second, we show that \texttt{\ours} \textbf{uniformly boosts performance} on downstream robotics tasks -- i.e., $ \texttt{baseline} + \texttt{\ours}> \texttt{baseline}$ for every baseline representation considered! Thus, we conclude that the affordance information injected by our method is highly useful for robot learning, and (for now) cannot be learned in a purely self-supervised manner.\\

\begin{table}[t]
    \vspace{0.4cm}
    \centering
    \resizebox{\linewidth}{!}{%

        \begin{tabular}{lcccc}
            \toprule
             & \textbf{Teacher ResNet} & \multicolumn{3}{c}{\textbf{\ours Models}} \\
             Front Cam & 100DoH~\cite{100doh} & w/ Ego4D & w/ ImageNet & w/ CLIP \\
             \midrule
             \textit{Toasting} & $35\% \pm 15\%$ & $\textbf{83\%} \pm 9\%$  & $75\% \pm 10\%$ & $50\% \pm 11\%$\\
             \textit{Pouring} & $34\% \pm 13\%$ & $\textbf{60\%} \pm 11\%$  & $48\% \pm 12\%$ & $39\% \pm 11\%$\\
             \textit{Stacking} & $0\%$ & $\textbf{77\%} \pm 10\%$  & $70\% \pm 11\%$ & $57\% \pm 11\%$\\
             \midrule
             \textbf{Average} & $35\% \pm 10\%$ & $\textbf{73\%} \pm 6\%$  & $64\% \pm 7\%$ & $48\% \pm 6\%$\\
            \bottomrule
        \end{tabular}
    }
    
    \caption{ \small This table compares 3 representations trained w/ \texttt{\ours} against the teacher ResNet~\cite{100doh} that generated our human affordance dataset (see Sec.~\ref{sec:prelims-affordances}). We find that the ResNet teacher under-performs even the worst \texttt{\ours} representation (fine-tuned from \texttt{CLIP}), \textit{even after excluding the stacking task, which it failed on.} }
    \vspace{-0.5cm}
    \label{tab:resnet-ablation}
\end{table}

\para{Second Camera View} A common critique is that robotic representations perform very differently when the camera view (even slightly) changes. To address this issue, we replicated the first experiment using a radically different \textit{ego view}, where the camera is placed over the robot's shoulder (i.e., on its ``head"). While perhaps a more realistic view, it is significantly more challenging due to the increased robot-object occlusion. Average success rates are presented in Fig.~\ref{fig:main} (right), and a per-task breakdown is in Appendix~\ref{app:ego-full-breakdown}. 
Note that our findings replicate almost exactly from the front camera view. The ImageNet MAE representation is still competitive with the other baselines, and applying \texttt{\ours} uniformly improves the baseline performance. In addition, we find that \textbf{\ours injects a higher level of robustness to camera view shifts}, when compared to the baselines. For example, we find that \texttt{ImageNet + \ours} performs the same on the ego and front camera, even though the \texttt{ImageNet} baseline clearly prefers the front cam. This general effect holds (to varying degrees) across all six baselines!\\

\begin{table*}[t]
    \vspace{0.2cm}
    \centering
    \resizebox{0.8\linewidth}{!}{%

        \begin{tabular}{lcccccc}
            \toprule
             & \multicolumn{2}{c}{\textbf{Ego4D}} & \multicolumn{2}{c}{\textbf{ImageNet}} & \multicolumn{2}{c}{\textbf{CLIP}} \\
             & + \ours & + Semantic &  + \ours & + Semantic  & + \ours &+ Semantic\\
             \midrule
             \textit{Toasting} & $\textbf{83\%} \pm 9\%$ & $25\% \pm 13\%$  & $\textbf{75}\% \pm 10\%$ & $40\% \pm 14\%$ & $\textbf{50\%} \pm 11\%$ & $20\% \pm 13\%$\\
             \textit{Pouring}  & $\textbf{60\%} \pm 11\%$ & $30\% \pm 13.4\%$ & $\textbf{48\%} \pm 12\%$ & $26\% \pm 11\%$ & $\textbf{39\%} \pm 11\%$ & $22\% \pm 10\%$\\
             \textit{Stacking} & $\textbf{77\%} \pm 10\%$ & $30\% \pm 11\%$  & $\textbf{70\%} \pm 11\%$ & $40\% \pm 12\%$ & $\textbf{57\%} \pm 11\%$ & $30\% \pm 13\%$\\
             \midrule 
             \textbf{Average} & $\textbf{73\%} \pm 6\%$ & $28\% \pm 7\%$ & $\textbf{64\%} \pm 7\%$ & $35\% \pm 7\%$ & $\textbf{48\%} \pm 6\%$ & $24\% \pm 7\%$\\
            \bottomrule
        \end{tabular} 
    }
    
    \caption{\small We create \texttt{Semantic} representations by fine-tuning the \texttt{Ego4D}, \texttt{ImageNet}, and \texttt{CLIP} baselines using a classification loss, instead of \ours's affordance loss. Note that the exact same Ego4D clips (see Sec.~\ref{sec:prelims-affordances}) are used during semantic fine-tuning, thanks to object class labels generated automatically by Detic~\cite{detic}. The sematic representations were evaluated (using the same BC pipeline) on the Toasting, Pouring, and Stacking tasks, and compared against their \texttt{HRP} counterparts. Success rates (and standard error) are reported above. We find that the affordance supervision provided by \ours is vastly superior to the semantic alternative. } 
    \vspace{-0.5cm}
    \label{tab:alt-sup}
\end{table*}

\para{Scaling to More Robots} Finally, we verify that \texttt{\ours} representations can provide benefits on other robotic hardware setups. Specifically, we compare \texttt{Ego4D + \ours} and \texttt{ImageNet + \ours} versus the respective baselines on the \textit{Pot on Stove} (xARM) and \textit{Hand Lift Cup} (dexterous hand) tasks. Results are presented in Table~\ref{tab:extra-robot}. Note that \texttt{\ours} representations provide consistent and significant performance during policy learning on these radically different robot setups, which both also use a unique side camera view. This gives us further confidence in \ours's view robustness and demonstrates that these representations are not tied to specific hardware setups, and can scale to complex morphologies like dexterous hands. 

\subsection{Distillation w/ \ours Improves Over Label Networks}
\label{sec:results-distill}

\begin{table}[t]
    \vspace{0.2cm}
    \centering
    \resizebox{\linewidth}{!}{%

        \begin{tabular}{lcccc}
            \toprule
             & \multicolumn{2}{c}{\textbf{Ego4D}}  & \multicolumn{2}{c}{\textbf{ImageNet}}\\
             & w/ \ours & Baseline & w/ \ours & Baseline \\
             \midrule
             \textit{Pot on Stove} & $\textbf{50\%} \pm 17\%$ & $40\% \pm 16\%$  & $\textbf{60\%} \pm 16\%$ & $40\% \pm 16\%$\\
             
             \textit{Hand Lift Cup} & $\textbf{50\%} \pm 17\%$ & $40\% \pm 16\%$  & $\textbf{50\%} \pm 17\%$ & $30\% \pm 15\%$\\
            \bottomrule
        \end{tabular}
    }
    
    \caption{\small We present results of \texttt{Ego4D + \ours} and \texttt{ImageNet + \ours}, as well as the respective baselines on the x-Arm (Pot on Stove) and a dexterous hand task (Lift Cup). We see that \texttt{\ours} can even boost performance in multiple morphologies, including a high-degree of freedom dexterous hand \cite{shaw2023Leaphand}. }
    \vspace{-0.5cm}
    \label{tab:extra-robot}
\end{table}


             
    

It is clear that applying \texttt{\ours} to self-supervised representations results in a consistent boost. However, the hand, object, and contact affordance labels for \texttt{\ours} themselves come from neural networks (see Sec.~\ref{sec:prelims-affordances}) -- specifically we use the ResNet-101~\cite{resnet} detector from 100DoH~\cite{100doh} as a label generator for our active object and contact affordance. The hand affordance we use comes from FrankMocap \cite{FrankMocap_2021_ICCV}, which uses  100DoH~\cite{100doh} as a base model. Thus, does distilling labels from this detector via \texttt{\ours} actually provide a benefit over simply using the 100DoH model itself as a pre-trained representation? To test this question, we fine-tune policies on the toasting, pouring, and stacking (front cam) tasks and compare them against \texttt{\ours} applied to \texttt{ImageNet}, \texttt{Ego4D}, and (the weakest model) \texttt{CLIP} (see Table~\ref{tab:resnet-ablation}). In all cases, our representation handily beats the 100DoH policy. So while the affordance labels can dramatically boost policy learning (via \texttt{\ours}), the source/teacher models are not at all competitive on robotics tasks.

\subsection{Comparing Against Alternate Forms of Supervision}
\label{sec:results-alternate}

\begin{table}[t]
    \vspace{0.3cm}
    \centering
    \resizebox{0.85\linewidth}{!}{%

        \begin{tabular}{lcc}
            \toprule
             Initialization & \textbf{w/ \ours} & \textbf{MAE Initialization} \\
             \midrule
             Ego4D & $\textbf{40\%}\pm 15\%$ & $15\% \pm 11\%$ \\
             ImageNet & $\textbf{40\%} \pm 15\%$ & $\textbf{40\%} \pm 15\%$ \\
            \bottomrule
        \end{tabular}
    }
    
    \caption{ \small This table compares \texttt{Ego4D + \ours} and \texttt{ImageNet + \ours} representations against their respective baselines on a \textit{stacking w/ distractors} task. Here the robot must successfully complete the usual stacking task, when extraneous objects (an orange carrot, and a green bowl) are added to the scene. We find that \texttt{Ego4D + \ours} improved over its baseline on this task, but \texttt{ImageNet + \ours} performed the same as its baseline. } 
    \vspace{-0.5cm}
    \label{tab:ood}
\end{table}

We now analyze if \ours's losses are better suited for robotics tasks than an alternate supervision scheme. To be clear, the previous results already demonstrated that \texttt{\ours + Ego4D} out-performed the \texttt{Ego4D} baseline by up to $20\%$ (see Fig.~\ref{fig:main}; left), despite being sourced from the same image data. However, it could be that the additional fine-tuning step with the $100K$ filtered interaction clips is responsible, and the specific affordance losses are not key. To test this, we ran a modified version of \ours using a semantic classification loss, instead of our affordance hand-object losses. The ground-truth labels for each image were obtained using the Detic object detector~\cite{detic}. We then similarly fine-tuned the \texttt{ImageNet}, \texttt{Ego4D}, and \texttt{CLIP} baseline representation using these labels, and compared them against the respective \texttt{\ours} models on the toasting, pouring, and stacking tasks. The results are presented in Table~\ref{tab:alt-sup} We find that the \texttt{\ours} models perform significantly better on every task. Thus, we conclude that \ours's affordance losses play an important role in boosting performance (i.e., it's not just data or extra fine-tuning).

\subsection{What Design Decisions are Important?}
\label{sec:results-ablate}

The following section ablates the key components of \texttt{\ours} to evaluate their relative importance. First, we apply \texttt{\ours} to each of the 6 baseline representations again, but this time none of the weights are kept fixed (see Sec.~\ref{sec:methods-implement}). These representations are fine-tuned on the toasting, stacking, and pouring tasks (front cam), and compared against the original \texttt{\ours} representations in Fig.~\ref{fig:ablate-nln}. Note that fine-tuning all the layers results in a substantial performance hit on average, and this trend is consistent regardless of the base representation! Thus, we conclude fine-tuning only the layer norms when applying \texttt{\ours} is the correct decision.

Next, we ablate each of the affordance losses in Eq.~\ref{eq:loss}, by applying \ours three times: once with $\lambda_\text{ct}=0$, then with $\lambda_\text{hand}=0$, and finally $\lambda_\text{obj}=0$. This process is repeated using 3 different base models; \texttt{ImageNet}, \texttt{Ego4D}, and \texttt{VC-1}. This creates 9 ablated models (3 losses x 3 initializations) that are compared versus the full \ours models on the toasting, pouring, and stacking tasks. The average results are presented in Fig.~\ref{fig:ablate-loss}, and the full, per-task breakdown is presented in the Appendix~\ref{app:ablation-breakdown}. 
We find that removing the object (Eq.~\ref{eq:loss-obj}) and hand (Eq.~\ref{eq:loss-hand}) losses uniformly results in significant performance degradation. Meanwhile, the contact loss (Eq.~\ref{eq:loss-ct}) only provides a significant boost for the \texttt{Ego4D} base model but does not affect the others. Thus, we conclude that object and hand losses are critical for our method, while the contact loss is more marginal, most likely due to the fact that the extraction of contacts is a relatively noisy process.

\begin{figure}[t]
    \centering
     \includegraphics[width=\linewidth]{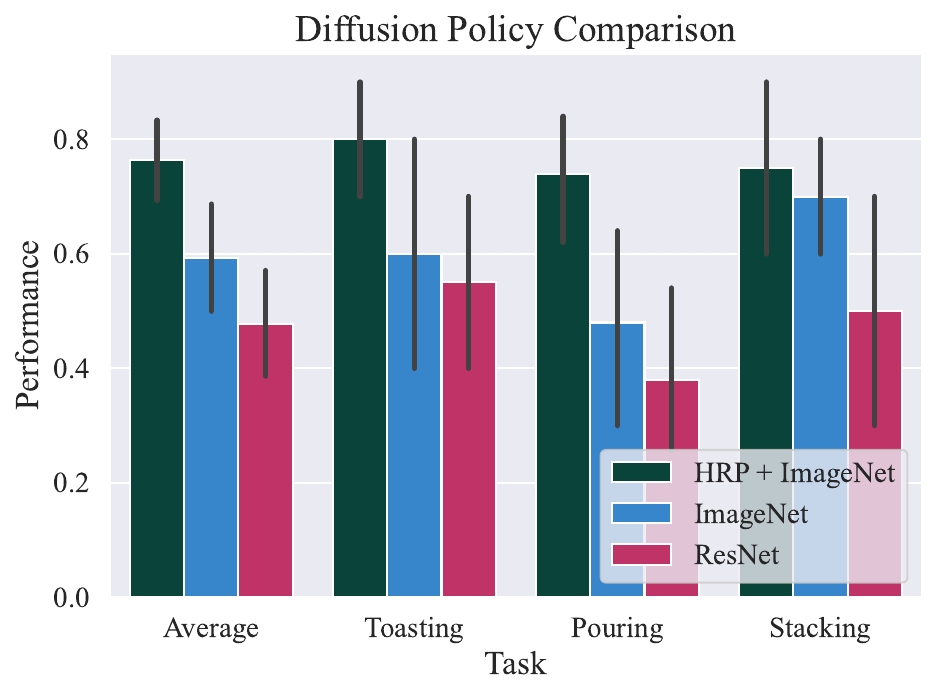}
     \caption{ \small This figure tests if \ours representations can boost performance when using a radically different imitation learning framework -- namely Diffusion Policy~\cite{chi2023diffusionpolicy}. We evaluate diffusion policies (following the U-Net + state action formula described by Chi et. al~\cite{chi2023diffusionpolicy}) on the toasting, pouring, and stacking tasks using 3 different visual encoders: the default ResNet encoder from RoboMimic~\cite{robomimic}, the \texttt{ImageNet + MAE} baseline, and our \texttt{\ours + ImageNet} features. We find a clear improvement when using \ours weights, which suggests that \ours is applicable to different imitation learning frameworks!}
     \label{fig:diffusion}
     \vspace{-0.3cm}
\end{figure}

\subsection{Novel Distractors During Test-Time}
\label{sec:results-distract}
We evaluate the performance of \texttt{\ours} and baseline approaches in OOD settings, by adding extraneous ``distractor" objects (an orange carrot and a light green bowl) in the stacking task. The robot must successfully ignore the distractor and complete the task. Results are presented in Table~\ref{tab:ood}. We found that both \texttt{ImageNet + \ours} and \texttt{ImageNet} had the same level of robustness to distractors. Meanwhile, \texttt{Ego4D}'s performance dropped substantially, while \texttt{Ego4D + \ours} remained robust. Our hypothesis is that human data by itself does not contain enough information to allow for OOD tasks. However, using \texttt{\ours} allows for more focus on task-relevant features, even when the representation is trained on less diverse data.

\subsection{Evaluating w/ Diffusion Policy}
\label{sec:results-diffusion}

Finally, we analyze if \texttt{\ours} representations offer improvements when using a radically different imitation learning framework, like diffusion policy~\cite{chi2023diffusionpolicy}. Specifically, we adopt the original U-Net action prediction head and environment setup from Chi et. al.~\cite{chi2023diffusionpolicy}, but replace their ResNet visual encoder (inspired from RoboMimic~\cite{robomimic}) with our \texttt{\ours + ImageNet} ViT-B model. Then we compare this \ours enhanced diffusion policy implementation, against (diffusion agents which use) both the original ResNet encoder and the baseline \texttt{ImageNet} ViT-B. Results for the (Franka) stacking, pouring, and toasting tasks are presented in Fig.~\ref{fig:diffusion}. We find that \texttt{HRP + ImageNet} significantly improves over both alternatives ($76\%$ for \ours v.s., $56\%$ for Chi et. al.'s implementation~\cite{chi2023diffusionpolicy}), despite using a radically different imitation learning objective/setup! Thus, we conclude that \ours representations can boost performance across different setups.

\section{Discussion and Future Work} 
\label{sec:discussion}



In this paper, we investigate human affordances as a strong prior for training visual representations. Thus, we present \ours, a semi-supervised pipeline that extracts contact points, hand poses, and activate objects from human videos, and uses these affordances for fine-tuning representations. \ours improves base model performance drastically, for five different, downstream behavior cloning tasks, across three robot morphologies and three camera views. All components of our approach, including LayerNorm tuning, our three affordances, and our distillation process (from affordance labels to representations) are important for the model's success. One key limitation of this approach is that it has only been tested on imitation settings in this paper. In the future, we hope to not only scale this approach to many more tasks and robot morphologies, but also incorporate \ours in other robot learning paradigms such as reinforcement learning or model based control.



\bibliographystyle{plainnat}
\bibliography{references}

\clearpage
\appendix
\subsection{Robot Controller Details}
\noindent \textbf{Franka}: We use a 7-DOF Franka Emika Panda robot arm with a parallel gripper, operating in delta end-effector action space. We use a VR-based teleoperation system to collect expert demos on Franka.

\noindent \textbf{xArm}: We use a 6-DOF xArm robot arm with a parallel gripper, operating in absolute end-effector action space. We use an off-the-shelf hand tracking system to collect expert demos on xArm.

\noindent \textbf{Dexterous Hand}: We use a 6-DOF xArm robot arm with a custom dexterous hand, operating in absolute end-effector space.

For each task, the expert gets to practice for 30 to 60 mins before collecting the demonstrations. We collect 50 expert demonstrations for each of the tasks.

\subsection{Front Cam: Full Task Performance Breakdown}
\label{app:front-full-breakdown}

\begin{table}[H]
\centering
\caption{Front Cam Performance Breakdown}
\label{my-label}
\resizebox{0.5\textwidth}{!}{%
\begin{tabular}{llcccc}
\toprule
\textbf{\makecell{Initial\\Representation}} & \textbf{Method} & \textbf{Toasting} & \textbf{Pouring} & \textbf{Stacking} & \textbf{Avg. (Real)} \\
\midrule
Ego4D & Baseline & 0.58 & 0.36 & 0.60 & 0.51 \\
 & Ours & \textbf{0.83} & \textbf{0.60} & \textbf{0.77} & \textbf{0.73} \\
\cmidrule{1-6}
ImageNet & Baseline & 0.53 & 0.45 & 0.47 & 0.48 \\
 & Ours & \textbf{0.75} & \textbf{0.48} & \textbf{0.70} & \textbf{0.64} \\
\cmidrule{1-6}
CLIP & Baseline & 0.28 & 0.33 & 0.33 & 0.32 \\
 & Ours & \textbf{0.50} & \textbf{0.39} & \textbf{0.57} & \textbf{0.48} \\
\cmidrule{1-6}
DINO & Baseline & 0.38 & 0.32 & 0.40 & 0.37 \\
 & Ours & \textbf{0.67} & \textbf{0.57} & \textbf{0.50} & \textbf{0.58} \\
\cmidrule{1-6}
MVP & Baseline & 0.27 & 0.41 & 0.47 & 0.38 \\
 & Ours & \textbf{0.73} & \textbf{0.44} & \textbf{0.63} & \textbf{0.60} \\
\cmidrule{1-6}
VC1 & Baseline & 0.52 & 0.33 & 0.57 & 0.47 \\
 & Ours & \textbf{0.83} & \textbf{0.34} & \textbf{0.53} & \textbf{0.57} \\
\bottomrule
\end{tabular}
}
\end{table}

We observe that HRP (Ours) consistently boosts the performance across all three tasks for the front cam.







\subsection{Ego Cam: Full Task Performance Breakdown}
\label{app:ego-full-breakdown}

\begin{table}[H]
\centering
\caption{Ego Cam Performance Breakdown}
\label{my-label}
\resizebox{0.5\textwidth}{!}{%
\begin{tabular}{llcccc}
\toprule
\textbf{\makecell{Initial\\Representation}} & \textbf{Method} & \textbf{Toasting} & \textbf{Pouring} & \textbf{Stacking} & \textbf{Avg. (Real)} \\
\midrule
Ego4D & Baseline & 0.2 & 0.12 & 0.3 & 0.21 \\
 & Ours & 0.2 & \textbf{0.22} & \textbf{0.45} & \textbf{0.29} \\
\cmidrule{1-6}
ImageNet & Baseline & 0.3 & 0.3 & 0.45 & 0.35 \\
 & Ours & \textbf{0.6} & \textbf{0.48} & \textbf{0.7} & \textbf{0.59} \\
\cmidrule{1-6}
CLIP & Baseline & 0.2 & 0 & 0 & 0.07 \\
 & Ours & \textbf{0.275} & \textbf{0.02} & 0 & \textbf{0.1} \\
\cmidrule{1-6}
DINO & Baseline & 0.35 & 0.32 & 0.3 & 0.32 \\
 & Ours & \textbf{0.45} & \textbf{0.7} & \textbf{0.55} & \textbf{0.57} \\
\cmidrule{1-6}
MVP & Baseline & 0.175 & 0.32 & 0.45 & 0.32 \\
 & Ours & \textbf{0.3} & \textbf{0.4} & \textbf{0.65} & \textbf{0.45} \\
\cmidrule{1-6}
VC1 & Baseline & 0.5 & 0.28 & 0.4 & 0.39 \\
 & Ours & \textbf{0.55} & \textbf{0.6} & \textbf{0.65} & \textbf{0.6} \\
\bottomrule
\end{tabular}
}
\end{table}

We also find that HRP (Ours) consistently boosts the performance across all three tasks for the ego camera. 

\subsection{Ablation Breakdown}
\label{app:ablation-breakdown}



\begin{table}[H]
\centering
\caption{Fine-Tuning Ablation Breakdown}
\label{tab:supp-ftablate}
\resizebox{0.5\textwidth}{!}{%
\begin{tabular}{llcccc}
\toprule
\textbf{\makecell{Initial\\Representation}} & \textbf{\makecell{Finetuning\\Scheme}} & \textbf{Toasting} & \textbf{Pouring} & \textbf{Stacking} & \textbf{Avg. (Real)} \\
\midrule
Ego4D & All Weights & \textbf{0.92} & 0.51 & 0.77 & 0.73 \\
 & LayerNorm (Ours) & 0.83 & \textbf{0.60} & 0.77 & 0.73 \\
\cmidrule{1-6}
ImageNet & All Weights & \textbf{0.82} & 0.34 & 0.63 & 0.60 \\
 & LayerNorm (Ours) & 0.75 & \textbf{0.48} & \textbf{0.70} & \textbf{0.64} \\
\cmidrule{1-6}
CLIP & All Weights & 0.23 & 0.27 & 0.13 & 0.21 \\
 & LayerNorm (Ours) & \textbf{0.50} & \textbf{0.39} & \textbf{0.57} & \textbf{0.48} \\
\cmidrule{1-6}
DINO & All Weights & 0.57 & 0.39 & 0.40 & 0.45 \\
 & LayerNorm (Ours) & \textbf{0.67} & \textbf{0.57} & \textbf{0.50} & \textbf{0.58} \\
\cmidrule{1-6}
MVP & All Weights & 0.45 & 0.39 & 0.47 & 0.43 \\
 & LayerNorm (Ours) & \textbf{0.73} & \textbf{0.44} & \textbf{0.63} & \textbf{0.60} \\
\cmidrule{1-6}
VC1 & All Weights & 0.52 & 0.41 & 0.47 & 0.47 \\
 & LayerNorm (Ours) & \textbf{0.83} & \textbf{0.34} & \textbf{0.53} & \textbf{0.57} \\
\bottomrule
\end{tabular}
}
\end{table}

\begin{table}[H]
\centering
\caption{Loss Ablation Performance Breakdown}
\label{tab:supp-ablate}
\resizebox{0.5\textwidth}{!}{%
\begin{tabular}{llcccc}
\toprule
\textbf{\makecell{Initial\\Representation}} & \textbf{Condition} & \textbf{Toasting} & \textbf{Pouring} & \textbf{Stacking} & \textbf{\makecell{Avg.\\(Real)}} \\
\midrule
Ego4D & No Contact & 0.65 & 0.34 & 0.5 & 0.50 \\
 & No Object & 0.425 & 0.42 & 0.3 & 0.38 \\
 & No Hand & 0.625 & 0.48 & 0.4 & 0.50 \\
 & Ours & \textbf{0.9} & \textbf{0.66} & \textbf{0.75} & \textbf{0.77} \\
\cmidrule{1-6}
Imagenet & No Contact & 0.625 & \textbf{0.64} & 0.7 & 0.66 \\
 & No Object & 0.525 & 0.52 & 0.55 & 0.53 \\
 & No Hand & 0.525 & 0.3 & \textbf{0.7} & 0.51 \\
 & Ours & \textbf{0.8} & 0.62 & \textbf{0.7} & \textbf{0.71} \\
\cmidrule{1-6}
VC-1 & No Contact & \textbf{0.625} & \textbf{0.48} & 0.75 & \textbf{0.62} \\
 & No Object & 0.225 & 0.38 & 0.65 & 0.42 \\
 & No Hand & 0.5 & 0.44 & 0.4 & 0.45 \\
 & Ours & 0.525 & 0.44 & \textbf{0.8} & 0.59 \\
\bottomrule
\end{tabular}
}
\end{table}

\textbf{Note:} do not compare numbers between Table~\ref{tab:supp-ablate} and the other tables. The loss ablation experiments were run on a separate day, so all numbers were re-ran on that day. This was done to ensure a proper A/B comparison between the all methods in this table. 





\subsection{Loss Weighting Sweep} 
\label{app:hparam}
We swept through a range of weights for each of the losses to narrow down on a particular set of loss weights for HRP (presented in Table~\ref{tab:loss-sweep}). These were based on relative orders of magnitude of the ground truth labels in the dataset. We empirically saw that increasing the loss weights by more than 0.5 negatively affected performance and led to collapse.

\begin{table}[H]
    \caption{\small We present the different affordance loss weights we ran sweeps on.}
    \centering
    \resizebox{\linewidth}{!}{%
        \begin{tabular}{lcc}
            \toprule
             Exp & \textbf{Loss Weights} \\
             \midrule
             \ours & $\lambda_{obj}=0.05$, $\lambda_{ct}=0.005$, $\lambda_{hand}=0.5$  \\
             Drop Contact Only & $\lambda_{obj}=0.05$, $\lambda_{ct}=0$, $\lambda_{hand}=0.5$  \\
             Drop Object Only & $\lambda_{obj}=0$, $\lambda_{ct}=0.005$, $\lambda_{hand}=0.5$  \\
             Drop Hand Only & $\lambda_{obj}=0.05$, $\lambda_{ct}=0.005$, $\lambda_{hand}=0$  \\
            \bottomrule
        \end{tabular}
    }
    \vspace{-0.1cm}
 
    \label{tab:loss-sweep}
\end{table}

\subsection{Data Pipeline Description}
\label{app:data_pipeline}
To obtain human data, we first extract video clips from Ego4D \cite{ego4d}. Our dataset contains approximately 1200 videos. Each video is broken down semantically into smaller clips by human annotators (as a part of the Ego4D). Our clips are between 1 and 30 seconds. For a given clip, we pass every frame through the 100 DOH model \cite{100doh}, which gives us hand object contact information. These are $\{h_l, h_r, o_l, o_r, c_l, c_r\}$. $h$ are the hand bounding boxes, $o$ are the object bounding boxes (which are in contact with the hand). $c$ are contact variable (i.e. fixed, portable, self or no contact). We only look at contacts with fixed and portable. $r$ or $l$ represents the left or right hand. Active object and hand trajectories used for our representations are directly used. For contact points, it is assumed that at the start of the clip there is no contact, from where we find the frame of first contact t. Since per-frame predictions are noisy, we run a filter \cite{savitzky1964smoothing} over the predictions. From the contact frame, we obtain the hand-bounding box $h$ and object bounding $o$. Contact points are computed in the intersection of $h$ and $o$, and the exterior of the hand. This exterior is obtained via skin segmentation (similar to \cite{bahl2023affordances, hoi}. These contacts can then be projected to previous frames in the clips by the homography matrix $H_t$ obtained via SIFT features.


\subsection{Behavior Cloning Hyper-Parameters}
\label{app:bc_params}

We list the hyper-paramaters that we used for policy training using behavior-cloning in this section. As shown in Figure~\ref{fig:policy-training}, we pass an image through the learned HRP visual representation to obtain a 768-dimensional latent vector. This latent vector is passed through a two-layer MLP with (512, 512) hidden layer dimensions. To the output of the MLP we apply RELU activation along with dropout regularization with prob=0.2 to estimate the mean ($\mu$), the mixing parameters ($\phi$), and the standard deviation ($\sigma$) of a Gaussian Mixture Model (GMM) distribution with 5 modes.

We choose GMM model based on prior work \cite{robomimic} that showed its crucial role in increasing BC performance. We use ADAM optimizer \cite{kingma2014adam} with the learning rate set to 1e-4, l2 weight decay also set to 1e-4. We train policy for 50K iterations. We also apply data augmentation (random crop and random blur) for the input images. We use the same set of hyper-parameters for both the real-world and the simulation tasks.

\subsection{Simulation Results}
\label{app:sim_performance}

For simulation tasks, we choose 5 tasks from the \textit{Metaworld}~\cite{metaworld} benchmark namely: BinPick, ButtonPress, Hammering, Drawer Opening, and Assembly. This benchmark is extensively used by the robot learning community. We used the same camera viewpoint, object sets, and expert demonstrations as used by prior work ~\cite{vc1}. We report the average performance on all 5 tasks in table ~\ref{tab:sim-perf}.

\begin{table}[H]
\centering
\caption{Sim Performance}
\label{tab:sim-perf}
\resizebox{0.5\textwidth}{!}{%
 \begin{tabular}{llc}
\toprule
\textbf{\makecell{Initial\\Representation}} & \textbf{Method} & \textbf{MetaWorld Avg Performance}\\
\midrule
Ego4D & Baseline & \textbf{0.656} \\
      & Ours     & 0.580 \\
\cmidrule{1-3}
ImageNet & Baseline & 0.556 \\
         & Ours     & \textbf{0.664} \\
\cmidrule{1-3}
CLIP & Baseline & \textbf{0.444} \\
     & Ours     & 0.408 \\
\cmidrule{1-3}
DINO & Baseline & 0.660 \\
     & Ours     & \textbf{0.664} \\
\cmidrule{1-3}
MVP & Baseline & 0.592 \\
    & Ours     & \textbf{0.640} \\
\cmidrule{1-3}
VC1 & Baseline & 0.576 \\
    & Ours     & \textbf{0.648} \\
\bottomrule
\end{tabular}}

\end{table}

\subsection{Evaluation}
\label{app:eval}
For each task, we run around 50 trials (per model), at various initial poses (for objects) and with different variations in objects. In every task, about half the trials are from the training distribution and half are from the test. The differences in objects include different colors, shapes, and even semantic differences: for example in the toasting task, plush toys were tested instead of the vegetables used to train. Cups or bowls were tested, instead of mugs that were used to train the pouring task, etc. 

Lighting is not controlled between train and test. We did try to run all baselines and methods as closely together as possible to avoid any confounding factors: i.e. for every trial, we ran all the baselines and our method together. Across trials, we allowed for variation in lighting conditions.

The results presented in the paper are the average of the successes, on a scale from 0 to 1. We present the criteria for success in each task: 

\begin{itemize}
    \item \textbf{Stacking:} 1 if the robot correctly picks and stacks the red block, and 0.5 if the red block is unstably placed on the green block.  
    \item \textbf{Pouring:} The fraction of candies, out of 5, successfully poured (e.g., $2 / 5$ candies poured $\rightarrow 0.4$ success). 
    \item \textbf{Toasting:} 1 if the whole task is completed, and 0.5 successful if the robot only successfully places the object.
    \item \textbf{Pot on Stove:} 1 if the food is correctly placed in the pot.
    \item \textbf{Hand Lift Cup} 1 if the cup is stably grasped and picked.
    
\end{itemize}

We also compute the standard error for these trials and show that as our confidence in Tables 1-3, and as an error bar in Figures 5-7.

\end{document}